\PassOptionsToPackage{unicode}{hyperref}
\PassOptionsToPackage{hyphens}{url}
\documentclass[
]{article}
\usepackage{xcolor}
\usepackage{amsmath,amssymb}
\usepackage{iftex}
\ifPDFTeX
  \usepackage[T1]{fontenc}
  \usepackage[utf8]{inputenc}
  \usepackage{textcomp} 
\else 
  \usepackage{unicode-math} 
  \defaultfontfeatures{Scale=MatchLowercase}
  \defaultfontfeatures[\rmfamily]{Ligatures=TeX,Scale=1}
\fi
\usepackage{lmodern}
\ifPDFTeX\else
\fi
\IfFileExists{upquote.sty}{\usepackage{upquote}}{}
\IfFileExists{microtype.sty}{
  \usepackage[]{microtype}
  \UseMicrotypeSet[protrusion]{basicmath} 
}{}
\makeatletter
\@ifundefined{KOMAClassName}{
  \IfFileExists{parskip.sty}{%
    \usepackage{parskip}
  }{
    \setlength{\parindent}{0pt}
    \setlength{\parskip}{6pt plus 2pt minus 1pt}}
}{
  \KOMAoptions{parskip=half}}
\makeatother
\usepackage{graphicx}
\usepackage{tikz}
\usetikzlibrary{calc}
\makeatletter
\newsavebox\pandoc@box
\newcommand*\pandocbounded[1]{
  \sbox\pandoc@box{#1}%
  \Gscale@div\@tempa{\textheight}{\dimexpr\ht\pandoc@box+\dp\pandoc@box\relax}%
  \Gscale@div\@tempb{\linewidth}{\wd\pandoc@box}%
  \ifdim\@tempb\p@<\@tempa\p@\let\@tempa\@tempb\fi
  \ifdim\@tempa\p@<\p@\scalebox{\@tempa}{\usebox\pandoc@box}%
  \else\usebox{\pandoc@box}%
  \fi%
}
\def\fps@figure{htbp}
\makeatother
\ifLuaTeX
  \usepackage{luacolor}
  \usepackage[soul]{lua-ul}
\else
  \usepackage{soul}
\fi
\usepackage{bookmark}
\IfFileExists{xurl.sty}{\usepackage{xurl}}{} 
\hypersetup{
  hidelinks,
  pdfcreator={LaTeX via pandoc}}
\makeatletter
\g@addto@macro\UrlBreaks{\do\0\do\1\do\2\do\3\do\4\do\5\do\6\do\7\do\8\do\9}
\makeatother

\title{Hybrid E-Assessment in Higher Education:\\
Semi-Automated Grading of Paper-Based Written Examinations}
\author{Hartwig Grabowski\thanks{Institute for Machine Learning and Analytics, Hochschule Offenburg, Offenburg, Germany. Email: hartwig.grabowski@hs-offenburg.de. ORCID: 0009-0001-4300-2626.}
\and
Michael Canz\thanks{Hochschule Offenburg, Offenburg, Germany. Email: michael.canz@hs-offenburg.de. ORCID: 0009-0008-5498-4814.}}
\date{}

\begin{document}

\maketitle

\begin{abstract}
This paper examines the limitations of fully digital and partially
digital e-assessment approaches in summative examinations in higher
education. The analysis focuses on the didactic narrowing caused by
closed question formats and on organizational, technical, and legal
constraints that become particularly relevant in large student cohorts.
As an alternative, the paper proposes a hybrid e-assessment approach that
retains paper-based, problem-oriented examination tasks while enabling
semi-automated grading. Assessment-relevant intermediate results are
encoded in a structured answer format, entered by students by hand, and
subsequently captured from table fields. The central technical bottleneck
is reliable recognition of handwritten characters under realistic
examination conditions. Recent vision-capable large language models,
combined with a two-pass validation principle and comparison against a
solution key, can reduce misclassifications and thereby improve the
validity, fairness, and scalability of summative assessment.
\end{abstract}

\textbf{Keywords:} hybrid e-assessment; summative examinations;
semi-automated grading; handwriting recognition; large language models

\section{Introduction}\label{introduction}

The transition from traditional paper-based examinations to digitally
supported assessment formats is not a simple change of medium. It
requires a careful redesign of established assumptions about examinations
and their role in higher education (Schmid et al., 2017). Digital
assessment promises efficiency, consistency, and new forms of processing,
but it also creates constraints that shape which tasks can be designed,
administered, and graded in practice (Graf-Schlattmann et al., 2018).

E-assessments are examinations that use digital support in their
preparation, administration, grading, or evaluation. They may be
diagnostic, formative, or summative, and each type places different
demands on didactics, infrastructure, software, and legal governance.
Diagnostic assessments are used to determine an initial level of
competence, for example in language placement tests, and require suitable
coordination between educational goals, software tools, rooms, and
hardware (Schmees et al., 2013). Formative assessments accompany the
learning process and typically require less infrastructure, while the
connection between didactic intent and software design remains central
(Schmees et al., 2013). Summative assessments, by contrast, certify
learning outcomes at the end of a course or module. Because their results
affect academic progression and, indirectly, professional opportunities,
they must satisfy the same legal and procedural requirements as
traditional examinations (Forgó et al., 2016).

This paper focuses on summative e-assessment. It first discusses the
limitations of fully digital and partially digital approaches, especially
where examination tasks are expected to support complex reasoning and
problem solving. It then presents a hybrid approach that keeps the
didactic flexibility of paper-based examinations while using automated
character recognition to reduce grading workload.

\section{Summative E-Assessment in Higher Education}\label{summative-e-assessment-in-higher-education}

Summative assessments are conducted at the end of a learning process and
serve to evaluate learning progress or learning outcomes (Schmees et al.,
2013). They are among the most widely used assessment formats in higher
education. Digitally supported summative assessment can broadly be
divided into fully digital and partially digital approaches.

\subsection{Fully Digital Approaches}\label{fully-digital-approaches}

In fully digital approaches, the examination is taken entirely on a
computer. Students work on the tasks electronically, and the assessment
is also processed electronically. These systems offer the highest degree
of automation when closed question formats such as multiple-choice items
are used (Schmees et al., 2013; Schmid et al., 2017). Free-text answers,
however, cannot usually be graded fully automatically with the same level
of reliability. Systems such as ILIAS, JACK, LPLUS, and Q-Exam, which
were identified in the ``E-Assessment NRW'' project, therefore often
rely primarily on multiple-choice and single-choice questions.

According to the study by Riedel and Möbius (Riedel \& Möbius, 2018),
one of the main reasons why lecturers refrain from using digital
examinations is their doubt that existing task types can be transferred
into automatically gradable formats. In that study, 51.4\,\% of the
lecturers surveyed stated that the lack of representability of their
specific examination questions was a central obstacle. This skepticism
varies by discipline. While it affects approximately half of lecturers in
many fields, 76\,\% of respondents in mathematics and the natural
sciences reported only limited possibilities for automated transfer.

Three concerns are particularly relevant. First, complex levels of
competence are difficult to assess through simple programmed grading
logic. Tasks that involve transfer, reasoning, or problem solving require
differentiated evaluation, and lecturers are concerned that fully
automated grading may not do justice to the complexity of academic
performance. Second, technical input barriers arise whenever answers
contain material beyond ordinary alphabetic text. Mathematical formulae,
chemical symbols, sketches, diagrams, and derivations are difficult to
enter directly into many digital examination systems. Third, creating or
migrating high-quality item pools requires substantial initial effort,
especially when the tasks are intended to go beyond the recall of factual
knowledge.

\subsection{Partially Digital Approaches}\label{partially-digital-approaches}

Partially digital approaches include scan-based examinations. Students
write on paper, the completed scripts are digitized afterwards, and parts
of the grading workflow are then supported by software. Systems such as
EvaExam/evasys, which are familiar in many higher-education contexts,
support both online examinations and scanner-based paper examinations
(Ivanova et al., 2018). In the scan-based setting, automated evaluation
is usually limited to closed questions, while open responses still
require manual grading.

The key limitation lies in the structure of the examination. Such exams
often consist mainly of multiple-choice questions with predefined answer
boxes. Compared with a traditional written examination, this structure
can be cumbersome to design. The layout options are constrained by the
system: in many cases, each section allows only one question with its
corresponding response options. This leads to a low density of questions
per page and produces a page structure that differs substantially from a
conventional examination sheet. Figure~\ref{fig:evaexam} provides an
original schematic illustration of such a scan-based layout, similar in
principle to systems such as EvaExam/evasys.

\begin{figure}[htbp]
\centering
\resizebox{0.78\linewidth}{!}{%
\begin{tikzpicture}[x=1cm,y=1cm]
  \definecolor{lightgraybox}{RGB}{239,239,239}
  \definecolor{darkrule}{RGB}{90,90,90}
  \draw[thick] (0,0) rectangle (12,15.5);
  \draw[very thick] (0.25,15.25) -- (1.05,15.25);
  \draw[very thick] (0.25,15.25) -- (0.25,14.45);
  \draw[very thick] (11.75,15.25) -- (10.95,15.25);
  \draw[very thick] (11.75,15.25) -- (11.75,14.45);
  \draw[very thick] (0.25,0.25) -- (1.05,0.25);
  \draw[very thick] (0.25,0.25) -- (0.25,1.05);
  \draw[very thick] (11.75,0.25) -- (10.95,0.25);
  \draw[very thick] (11.75,0.25) -- (11.75,1.05);

  \node[anchor=west,font=\bfseries] at (0.65,14.75) {Scan-based examination sheet};
  \node[anchor=east,font=\small] at (11.35,14.75) {Page 1 of 6};
  \draw[darkrule] (0.6,14.4) -- (11.4,14.4);

  \node[anchor=west,font=\small] at (0.75,13.95) {Course: Introductory Mathematics};
  \node[anchor=west,font=\small] at (0.75,13.55) {Student ID:};
  \foreach \x in {0,...,8} {
    \draw (3.0+\x*0.45,13.3) rectangle +(0.32,0.32);
  }
  \node[anchor=west,font=\small] at (0.75,12.95) {Name: \rule{4.8cm}{0.3pt}};

  \draw[fill=lightgraybox,draw=darkrule] (0.7,12.25) rectangle (11.3,12.75);
  \node[anchor=west,font=\bfseries\small] at (0.95,12.5) {Section 1: Single-choice questions};

  \node[anchor=west,font=\small] at (0.95,11.9) {1.1\quad Where does \(x^2 - 1\) intersect the \(x\)-axis?};
  \foreach \x/\l in {1.2/{\(-1\) and \(1\)},3.6/{0},4.8/{1},6.0/{none}} {
    \draw (\x,11.45) rectangle +(0.24,0.24);
    \node[anchor=west,font=\scriptsize] at (\x+0.33,11.57) {\l};
  }
  \node[anchor=west,font=\small] at (0.95,11.05) {1.2\quad What is the slope of \(y=-\frac{1}{2}x+2\)?};
  \foreach \x/\l in {1.2/{\(-\frac{1}{2}\)},2.8/{2},4.0/{\(\frac{1}{2}\)},5.3/{\(-2\)}} {
    \draw (\x,10.6) rectangle +(0.24,0.24);
    \node[anchor=west,font=\scriptsize] at (\x+0.33,10.72) {\l};
  }
  \node[anchor=west,font=\small] at (0.95,10.2) {1.3\quad A non-zero polynomial of degree \(n\) has at most};
  \foreach \x/\l in {1.2/{\(n\) zeros},3.0/{\(n+1\)},4.4/{\(2n\)},5.8/{infinitely many}} {
      \draw (\x,9.75) rectangle +(0.24,0.24);
      \node[anchor=west,font=\scriptsize] at (\x+0.33,9.87) {\l};
  }

  \draw[fill=lightgraybox,draw=darkrule] (0.7,8.85) rectangle (11.3,9.35);
  \node[anchor=west,font=\bfseries\small] at (0.95,9.1) {Section 2: Formula recognition};
  \node[anchor=west,font=\small] at (0.95,8.25) {Select the derivative of \(f(x)=x^2-1\).};
  \foreach \x/\l in {1.15/{\(2x\)},3.0/{\(x\)},4.45/{\(x^2\)},5.95/{\(-1\)}} {
    \draw (\x,7.8) rectangle +(0.26,0.26);
    \node[anchor=west,font=\scriptsize] at (\x+0.35,7.93) {\l};
  }
  \node[anchor=west,font=\small] at (0.95,7.15) {Which expression is linear?};
  \foreach \x/\l in {1.15/{\(x^2+1\)},3.0/{\(-\frac{1}{2}x+2\)},5.1/{\(\frac{1}{x}\)},7.0/{\(2^x\)}} {
    \draw (\x,6.7) rectangle +(0.26,0.26);
    \node[anchor=west,font=\scriptsize] at (\x+0.35,6.83) {\l};
  }

  \draw[fill=lightgraybox,draw=darkrule] (0.7,5.85) rectangle (11.3,6.35);
  \node[anchor=west,font=\bfseries\small] at (0.95,6.1) {Section 3: Automatically graded statements};
  \node[anchor=west,font=\small] at (0.95,5.35) {Mark all true statements.};
  \foreach \y/\txt in {4.85/{A quadratic polynomial has at most two zeros.},4.25/{The slope of \(y=-\frac{1}{2}x+2\) is \(2\).},3.65/{The equation \(x^2-1=0\) has two real solutions.}} {
    \draw (1.05,\y-0.08) rectangle +(0.28,0.28);
    \node[anchor=west,font=\scriptsize] at (1.45,\y+0.05) {\txt};
  }

  \draw[fill=lightgraybox,draw=darkrule] (0.7,2.9) rectangle (11.3,3.4);
  \node[anchor=west,font=\bfseries\small] at (0.95,3.15) {Internal identification area};
  \node[anchor=west,font=\scriptsize] at (1.05,2.45) {Page and exam identifier};
  \draw[fill=black] (7.6,2.18) rectangle (10.9,2.72);
  \foreach \x in {7.85,8.25,8.9,9.5,10.25} {
    \draw[fill=white,draw=white] (\x,2.32) rectangle +(0.28,0.16);
  }
  \node[font=\scriptsize] at (9.25,1.95) {barcode for internal assignment};
\end{tikzpicture}%
}
\caption{Schematic illustration of a scan-based examination page with closed answer boxes, a student identifier area, scan markers, and an internal barcode for assigning pages to an examination. The figure is an original illustration created for this paper.}
\label{fig:evaexam}
\end{figure}

\section{Didactic Requirements for Summative E-Assessment}\label{didactic-requirements-for-summative-e-assessment}

The limited transferability of traditional written examination tasks
into automatically gradable formats points to a more fundamental
didactic problem. Summative examinations do not only test achievement;
they also shape learning (Crooks, 1988; Biggs, 1996; Black \& Wiliam,
1998). Students use past and expected examinations as
central learning material, especially during examination preparation, and
they often acquire additional understanding by actively engaging with the
tasks (Entwistle \& Entwistle, 1991; Roediger \& Karpicke, 2006).

This gives summative assessment a dual function. In addition to measuring
the achieved level of learning, examinations should support conceptual
consolidation. A high-quality written examination is therefore not
defined solely by valid performance measurement. It should also enable
students to engage in a reflective problem-solving process that can
contribute to durable learning (Gibbs \& Simpson, 2004). Consequently,
summative e-assessment must
be designed in a way that reconciles automated evaluation with
learning-oriented task formats.

\subsection{Example of Learning-Oriented Task Design}\label{example-of-learning-oriented-task-design}

A simple example from physics illustrates the point: conservation of
energy in free fall. Instead of asking students to recall the
independence of falling velocity from mass as isolated factual knowledge,
the examination task can be designed as a structured reasoning process:

\begin{itemize}
\item
  Derivation: students write the equation for potential energy at a
  given height, \(E_{pot} = m \cdot g \cdot h\).
\item
  Application: students formulate the equation for kinetic energy at
  impact, \(E_{kin} = \frac{1}{2} m v^{2}\).
\item
  Transfer: by equating both formulae, \(E_{pot} = E_{kin}\), students
  recognize within the calculation that the mass parameter \(m\)
  cancels.
\end{itemize}

The didactic value lies in the process. The examination becomes a
learning opportunity because students actively derive the physical
relationship. The insight that velocity does not depend on mass is the
result of a logical problem-solving process rather than merely recalled
knowledge.

\subsection{Limits of Fully Automated Evaluation}\label{limits-of-fully-automated-evaluation}

The demand for fully automated grading in summative e-assessment often
narrows the range of didactically meaningful task formats. Closed formats
such as multiple-choice and single-choice questions can represent
stepwise problem-solving processes only to a limited extent. Tasks based
on derivation, application, and transfer are therefore difficult to
implement adequately, especially when they require sketches, function
graphs, formula transformations, or complex visual representations.

Many fully digital systems also rely on HTML-based or otherwise
syntax-driven authoring processes. These processes can make it difficult
to design complex examination pages with the precision required in
summative assessment. As a result, the creation of examinations may
become time-consuming and error-prone. Highly automated systems are
therefore of limited use in contexts where the goal is not merely
efficient scoring but also a task architecture that supports learning and
problem solving.

\subsection{Organizational Constraints}\label{organizational-constraints}

In addition to didactic constraints, fully digital summative assessments
create substantial organizational and technical demands, particularly in
large cohorts. In a foundational course with, for example, 800 students,
a fully digital examination requires an equivalent number of computer
workstations. These devices must be configured consistently, tested in
advance, and coordinated so that all participants can start under stable
and comparable conditions. The operational reliability of the technical
environment becomes critical, since even isolated failures or system
crashes can disrupt the examination and may have legal consequences.

Further requirements concern invigilation and the physical examination
environment. Workstations may need privacy screens to reduce cheating
opportunities, and the examination system must be protected against
unauthorized access, manipulation, copying, and remote connections. Taken
together, these infrastructural, operational, and security-related
requirements create high entry barriers that often limit the practical
use of fully digital summative examinations despite their potential
automation benefits.

\subsection{Why Paper Still Matters}\label{why-paper-still-matters}

For didactically rich summative examinations, the established paper
format remains attractive. It minimizes organizational complexity and
allows the task architecture to be determined primarily by subject-matter
and didactic considerations rather than by technical format constraints.
This is especially important when examinations are intended not only to
measure performance but also to support reasoning and learning.

The implication is not that technology should be avoided. Rather,
technology should support grading and post-processing without constraining
the design of complex, stepwise problem-solving tasks. Paper provides the
necessary design freedom and is therefore a suitable starting point for
developing semi-automated grading methods that support, rather than
determine, the didactic requirements of summative assessment.

\section{Semi-Automated Assessment on Paper}\label{semi-automated-assessment-on-paper}

The approach proposed here combines the didactic flexibility of
paper-based written examinations with largely automated evaluation.
After an examination task has been designed, assessment-relevant
intermediate results are transferred into a structured gap or coding
format. Instead of being entered as free text, the relevant results are
recorded in predefined answer fields. Each gap is assigned a fixed set of
letter-based answer options, for example A to Z.

\begin{figure}[htbp]
\centering
\includegraphics[width=0.78\linewidth]{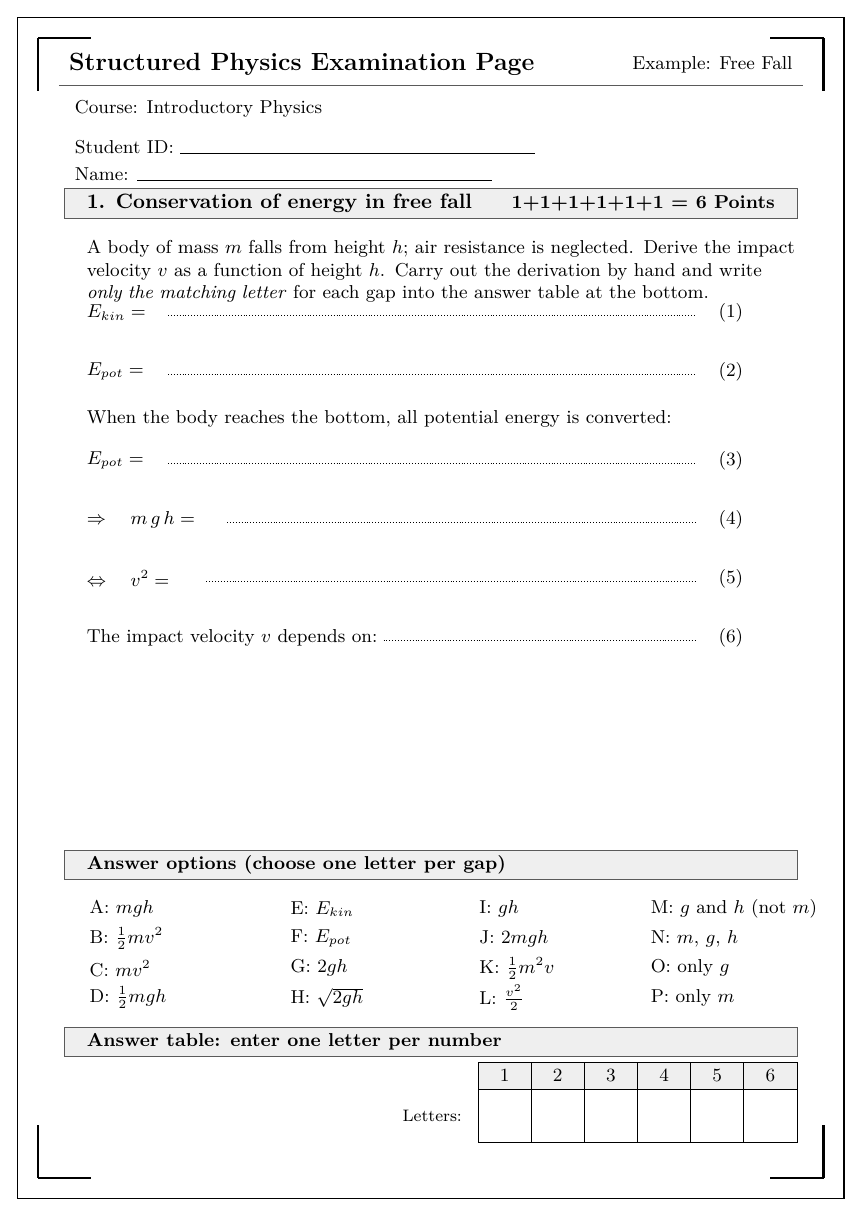}
\caption{Example of a structured examination page: the free-fall derivation from Section~\ref{example-of-learning-oriented-task-design} recast in the coding format. Students carry out the derivation by hand and transfer only the matching letter for each gap (1)--(6) into the answer table. The handwritten letters are captured automatically, while the open derivation documents the solution path. The large option pool keeps random guessing unlikely without turning the task into a closed question format.}
\label{fig:exam-page-main}
\end{figure}

The large number of predefined options reduces the likelihood of a
correct answer by random guessing without forcing the whole task into a
closed question format. Students still carry out the calculation,
derivation, or reasoning process by hand and only then transfer the
corresponding partial results into the designated fields. Automation
therefore begins at the standardized capture and evaluation of answers,
not at the restriction of the task architecture. Figure~\ref{fig:exam-page-main}
shows an example of such a structured examination page: the free-fall derivation
from Section~\ref{example-of-learning-oriented-task-design}, recast in the coding
format. Two further examples, a database query based on a schema diagram and a
program-completion task, are provided in Appendix~\ref{app:additional-examples}.

This decoupling is what separates the coding format from a checkbox
examination. In a closed multiple-choice item, such as the scan-based sheet in
Figure~\ref{fig:evaexam}, the answer options are the task: the question is
defined by its boxes, and the response is reduced to recognizing the correct
alternative among the few that are shown. The format itself pushes the item
toward recognition and away from derivation, which is precisely the narrowing
discussed in Section~\ref{limits-of-fully-automated-evaluation}.

In the coding format, by contrast, the letter assignment is only a thin capture
layer placed on top of a freely designed task. The didactic design of the
derivation is therefore independent of the scoring mechanism. This is the reason
the stepwise derivation survives: unlike the closed format, the coding scheme
does not force the task into a recognition shape, so the derivation can be
authored as in Figure~\ref{fig:exam-page-main} while still yielding a
machine-readable result.

The central challenge for largely automated evaluation is the reliable
machine recognition of the letters that students enter into the table
fields. These entries represent the encoded answers and therefore the
assessment-relevant intermediate results. The automation problem shifts
from interpreting freely formulated solution paths to robust character
classification under realistic examination conditions: variable
handwriting, different stroke widths, corrections, scanning artifacts,
and printing artifacts. For validity and fairness, recognition
performance must not only be high on average; misclassifications must be
systematically minimized, and uncertainty must be handled in a way that
does not cause unjustified loss of points.

\subsection{AI-Based Letter Recognition}\label{ai-based-letter-recognition}

An earlier implementation used a YOLO-5-based method (Redmon et al.,
2016) trained on image segments extracted from the relevant table fields
(Grabowski, 2023).
Despite this domain-specific training, recognition performance remained
at 88.28\,\%. For summative assessment, this level is insufficient:
roughly one error per ten recognized letters would create an
unacceptable risk of legally relevant misgrading. The result also shows
how difficult character classification remains under authentic
conditions, especially with variable handwriting, corrections, and scan
or print artifacts.

\begin{figure}[htbp]
\centering
\begin{minipage}[t]{0.58\linewidth}
\centering
\includegraphics[width=\linewidth]{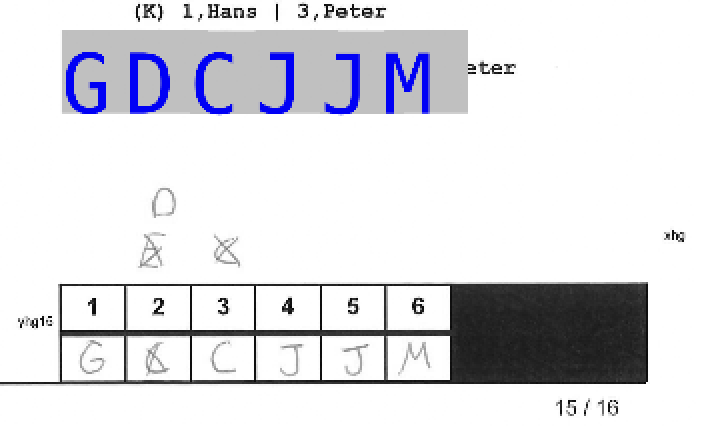}
\end{minipage}\hfill
\begin{minipage}[t]{0.30\linewidth}
\centering
\includegraphics[width=\linewidth]{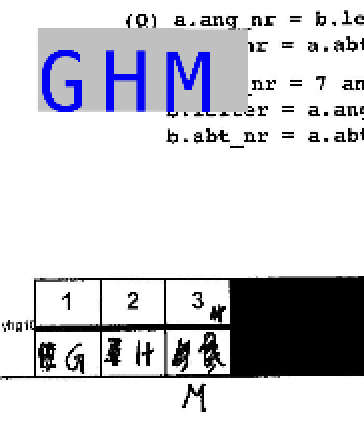}
\end{minipage}
\caption{Recognition is challenging when letters must be assigned to the correct columns (left) or when students cross out entries (right). The recognized letters are displayed above the table during processing.}
\label{fig:recognition-problems}
\end{figure}

Recent vision-capable large language models (LLMs) offer a more
promising technical path (Radford et al., 2021; Bommasani et al., 2021;
Bordes et al., 2024). Their advantage is not only that they classify
isolated glyphs. They can also consider the visual context of an entry
and its spatial position relative to the table structure. This makes it
possible to assign a handwritten letter more reliably to the intended
column even when the character is shifted, crosses cell boundaries, or is
placed atypically (Figure~\ref{fig:recognition-problems}).

\subsection{End-to-End Workflow}\label{end-to-end-workflow}

The complete workflow is more involved than isolated letter recognition.
After the examinations have been collected, the paper scripts are scanned
into a combined PDF document. Each relevant table is then digitally
cropped from the PDF and processed in two parallel LLM requests. In the
first request, the cropped examination material is provided together with
the correct solution, allowing the model to check whether the expected
letters appear in the corresponding table fields. In the second request,
the same image material is processed without the correct solution so that
the entered letters can be extracted independently.

\begin{figure}[htbp]
\centering
\includegraphics[width=\linewidth,height=0.66\textheight,keepaspectratio]{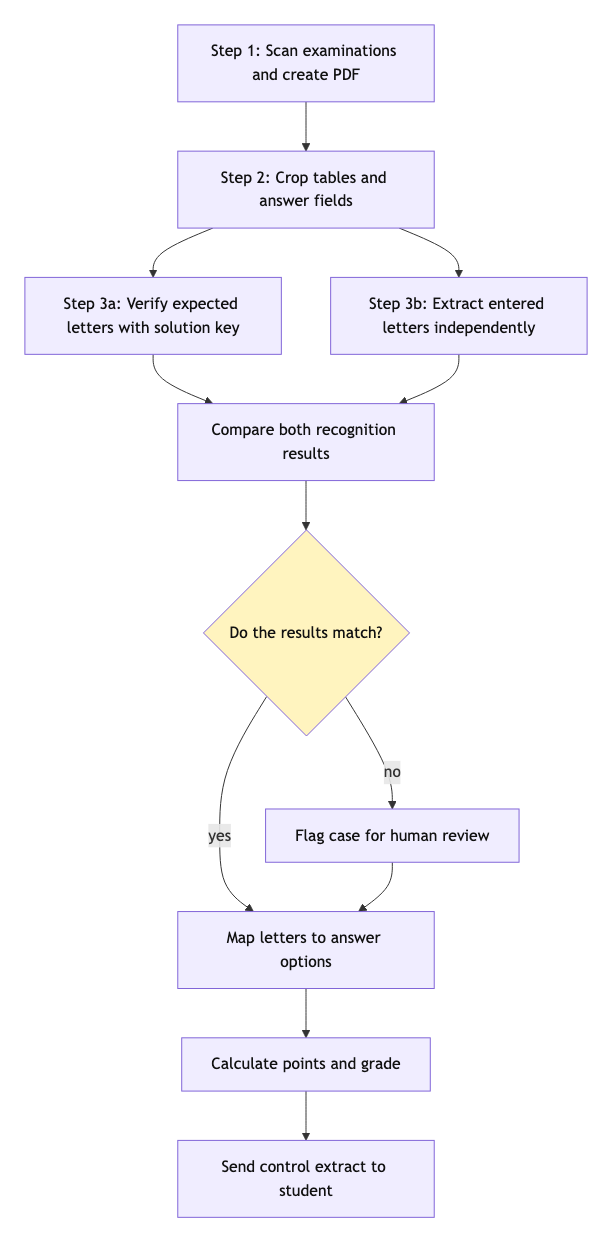}
\caption{The grading workflow can be decomposed into four stages. Using two LLM configurations allows the scanned solutions to be checked according to a two-pass validation principle.}
\label{fig:workflow}
\end{figure}

The results of both requests are compared. If they agree, the recognized
letters are mapped to the predefined answer options and linked to the
corresponding point values, enabling automatic calculation of total
points and grades. For additional validation, the automatically captured
student identifier can be used to send students an email with a cropped
image of the relevant table and the recognized letters. Students can then
check the recognition result and report deviations. The overall process
is summarized in Figure~\ref{fig:workflow}.

\subsection{Recognition Performance}\label{recognition-performance}

The recognition performance achieved with the newly evaluated vision
models was strong. In particular, Google Gemini 3 Flash showed the best
evaluation performance in the tested setting. Across 3009 letters to be
recognized, 35 incorrect recognitions occurred and required manual
post-processing (Table~\ref{tab:error-rate}).

\begin{table}[htbp]
\vspace*{0.8cm}
\centering
\begin{tabular}{lrr}
\hline
Model & Recognition & Errors \\
\hline
Gemini 3 Flash & \(\sim\)98.84\,\% & 35 \\
Qwen3-VL-30B-A3B-Instruct & \(\sim\)98.31\,\% & 51 \\
GPT-5 & \(\sim\)96.54\,\% & 104 \\
Grok 4.1 Fast & \(\sim\)88.90\,\% & 334 \\
\hline
\end{tabular}
\caption{Error rates of the evaluated language models in practical use. A total of 3009 letters were recognized and evaluated.}
\label{tab:error-rate}
\end{table}

The grading workload is reduced substantially because the automated
process handles the vast majority of entries reliably, while human review
can focus on a small number of clearly identifiable problem cases. This
makes large examination cohorts easier to manage without requiring a
complete transition to fully digital examination formats.

At the same time, the hybrid approach involves a non-trivial
exam-specific setup effort. Configuration files are needed to map pages
to table regions and table columns to assessment criteria and point
values. In recurring examination formats, existing configurations can be
reused and adjusted, but the initial setup remains a fixed cost for each
examination. The procedure is therefore less economical for small cohorts
or rarely repeated examinations. For large examination volumes, however,
the setup effort is typically amortized because the automated process
takes over most of the recognition work and concentrates manual effort on
a small number of uncertain cases.

\subsection{Application Scenarios}\label{application-scenarios}

Two application scenarios can be distinguished for practical use of the
hybrid e-assessment workflow.

In the first scenario, grading is performed page by page in an
interactive mode. The examiner moves through the examination
sequentially and triggers evaluation of the current page, for example by
using a shortcut. The recognized letters are then transferred
automatically into the corresponding table fields, after which the
examiner confirms the results visually or corrects them immediately
before moving to the next page. This approach provides maximum control at
page level but requires continuous examiner presence throughout the
grading process.

In the second scenario, evaluation is primarily conducted according to a
two-pass validation principle with two LLM configurations. Pages with
ambiguous recognition, especially where the two outputs diverge, are
flagged and reviewed in a final focused pass by the examiner.

In both scenarios, automated scoring, summation of points, and comparison
with the grading table follow the recognition step. The student
identifier can also be captured automatically from the scans to support
unambiguous assignment of results. For transparency and self-checking,
students receive only the digitized table excerpts and not the complete
examination script, avoiding broader release of examination documents.

\section{Conclusion}\label{conclusion}

The proposed hybrid method offers a pragmatic alternative to purely
computer-based summative assessment. It does not subordinate the examiner
or the task design to the technical system. Instead, it uses technical
support where it creates clear value: in reducing grading workload while
preserving the pedagogical role of written examinations.

The creation of examinations remains accessible because familiar word
processing tools and established paper-based workflows can still be
used. Existing paper examinations do not need to be rebuilt in
system-specific authoring tools. They can be adapted to a structured
answer-field or coding format with manageable effort, while maintaining
substantive and didactic freedom, especially for problem-oriented tasks.

At the same time, grading can be automated along the process described
above. Vision-capable LLMs support reliable recognition, and a two-pass
validation principle helps identify problem cases that require a
case-specific decision by the examiner. Overall, the approach accelerates
and simplifies grading, including the calculation of points and grades,
without reducing freedom in examination design.

\section*{References}\label{references}
\begingroup\sloppy\setlength{\parskip}{3pt plus 1pt minus 1pt}%

Biggs, J. (1996). Enhancing teaching through constructive alignment.
\emph{Higher Education}, 32(3), 347--364.
https://doi.org/10.1007/BF00138871

Black, P. \& Wiliam, D. (1998). Assessment and classroom learning.
\emph{Assessment in Education: Principles, Policy \& Practice}, 5(1),
7--74. https://doi.org/10.1080/0969595980050102

Bommasani, R., et al. (2021). On the opportunities and risks of
foundation models. \emph{arXiv preprint arXiv:2108.07258}.
https://doi.org/10.48550/arXiv.2108.07258

Bordes, F., et al. (2024). An introduction to vision-language modeling.
\emph{arXiv preprint arXiv:2405.17247}.
https://doi.org/10.48550/arXiv.2405.17247

Crooks, T. J. (1988). The impact of classroom evaluation practices on
students. \emph{Review of Educational Research}, 58(4), 438--481.
https://doi.org/10.3102/00346543058004438

Entwistle, N. J. \& Entwistle, A. (1991). Contrasting forms of
understanding for degree examinations: the student experience and its
implications. \emph{Higher Education}, 22(3), 205--227.
https://doi.org/10.1007/BF00132288

Forgó, N., Graupe, S. \& Pfeiffenbring, J. (2016). \emph{Rechtliche
Aspekte von E-Assessments an Hochschulen}.
https://doi.org/10.17185/DUEPUBLICO/42871

Gibbs, G. \& Simpson, C. (2004). Conditions under which assessment
supports students' learning. \emph{Learning and Teaching in Higher
Education}, 1, 3--31.

Grabowski, H. (2023). Intelligent character recognition of handwritten
forms with deep neural networks. In D. Cavallucci, P. Livotov, \& S.
Brad (Eds.), \emph{Towards AI-Aided Invention and Innovation} (Vol. 682,
pp. 81--94). Springer Nature Switzerland.
https://doi.org/10.1007/978-3-031-42532-5\_6

Graf-Schlattmann, M., Meister, D. M., Oevel, G. \& Wilde, M. (2018).
\emph{Hochschulstrategie als Prozess. Zum allgemeinen und
hochschulspezifischen Begriff der Strategie}.
https://doi.org/10.5281/ZENODO.1293797

Ivanova, M., Durcheva, M., Baneres, D. \& Rodriguez, M. E. (2018).
eAssessment by using a Trustworthy System in Blended and Online
Institutions. \emph{2018 17th International Conference on Information
Technology Based Higher Education and Training (ITHET)}, 1--7.
https://doi.org/10.1109/ITHET.2018.8424805

Radford, A., et al. (2021). Learning transferable visual models from
natural language supervision. \emph{Proceedings of the 38th International
Conference on Machine Learning (ICML)}, PMLR 139, 8748--8763.
https://doi.org/10.48550/arXiv.2103.00020

Redmon, J., Divvala, S., Girshick, R. \& Farhadi, A. (2016). You only
look once: Unified, real-time object detection. \emph{Proceedings of the
IEEE Conference on Computer Vision and Pattern Recognition (CVPR)},
779--788. https://doi.org/10.1109/CVPR.2016.91

Riedel, J. \& Möbius, K. (2018). Bestandsaufnahme, Hindernisse und
Möglichkeiten des Einsatzes von E-Assessment an sächsischen Hochschulen.
\emph{Beiträge zur Hochschulforschung}, 40(4), 68--86.

Roediger, H. L. \& Karpicke, J. D. (2006). Test-enhanced learning:
Taking memory tests improves long-term retention. \emph{Psychological
Science}, 17(3), 249--255.
https://doi.org/10.1111/j.1467-9280.2006.01693.x

Schmees, M., Krüger, M. \& Schaper, E. (2013). E-Assessments an
Hochschulen: Ein vielschichtiges Thema. In \emph{E-Assessments in der
Hochschullehre. Einführung, Positionen \& Einsatzbeispiele: Psychologie
und Gesellschaft} (No. 13, pp. 19--32). PL Academic Research\,:
Frankfurt, M. https://doi.org/10.25656/01:12879

Schmid, U., Goertz, L., Radomski, S., Thom, S., Behrens, J., \&
Bertelsmann Stiftung. (2017). \emph{Monitor Digitale Bildung: Die
Hochschulen im digitalen Zeitalter}. https://doi.org/10.11586/2017014
\par\endgroup

\clearpage
\appendix
\section{Additional Examples of Structured Examination Pages}\label{app:additional-examples}

\begin{center}
\fbox{\includegraphics[width=0.85\linewidth]{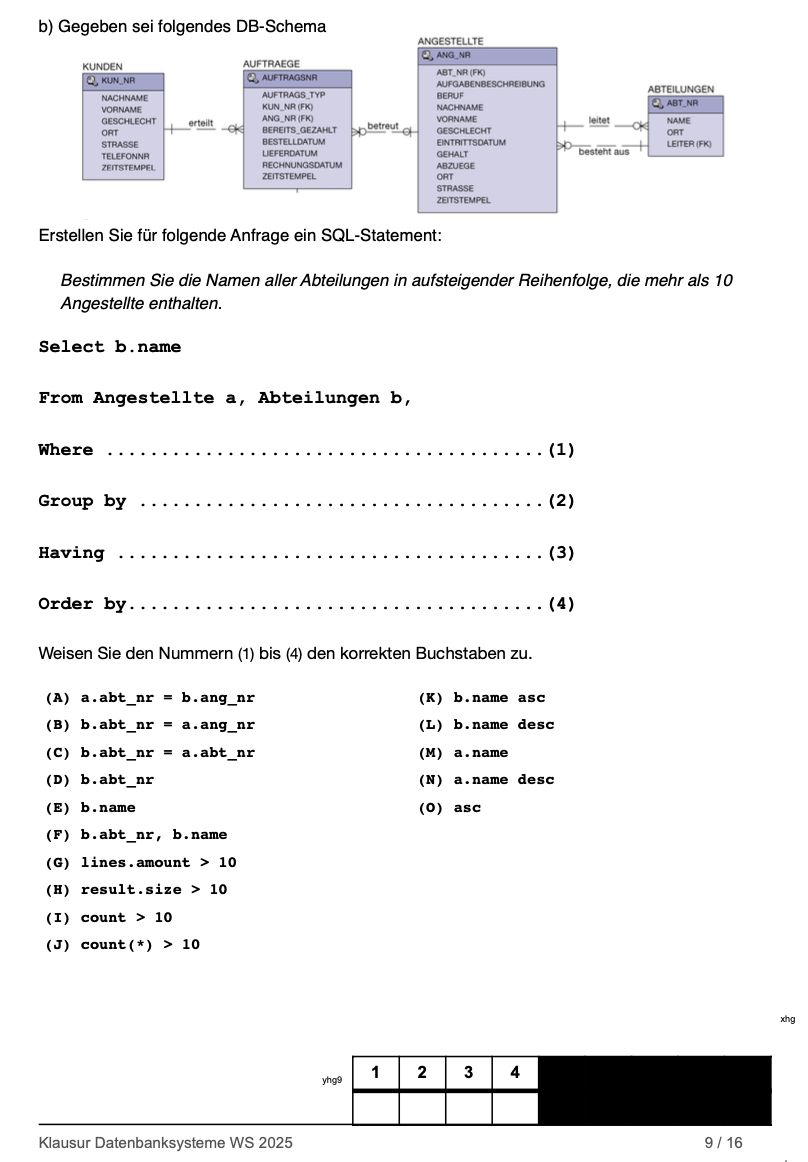}}\\[\baselineskip]
\refstepcounter{figure}
\label{fig:exam-page-sql}
\parbox{0.86\linewidth}{\small Figure~\thefigure: Structured examination page created in the usual way with a schema diagram and gap-based tasks. Each gap is assigned a unique letter option, and students enter the corresponding letters into the table at the bottom of the page. The page is an original German-language examination.}
\end{center}

\clearpage

\begin{center}
\fbox{\includegraphics[width=0.85\linewidth]{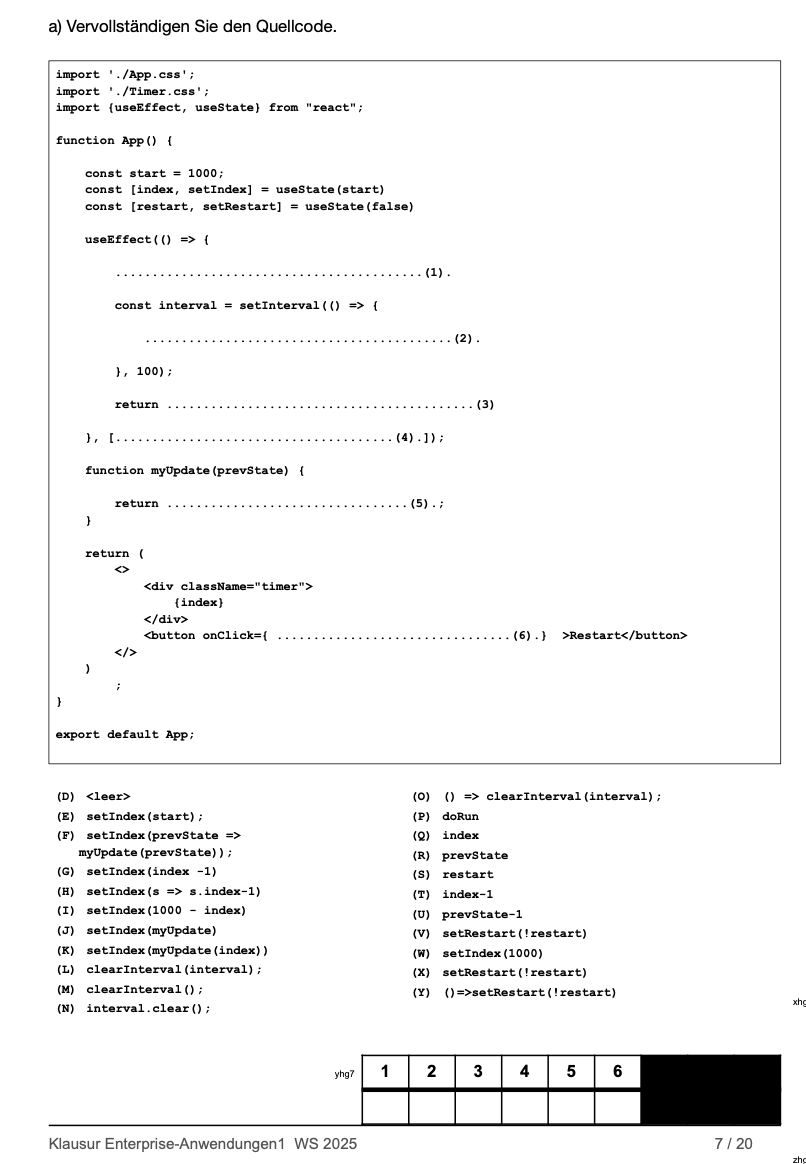}}\\[\baselineskip]
\refstepcounter{figure}
\label{fig:program-completion-example}
\parbox{0.86\linewidth}{\small Figure~\thefigure: Additional example of a structured examination page using a program-completion task. The same coding principle can be applied beyond mathematical derivations and diagram-based tasks. The page is an original German-language examination.}
\end{center}

\end{document}